\pdfoutput=1

\documentclass[11pt]{article}

\usepackage[]{naacl2021}

\usepackage{times}
\usepackage{latexsym}
\usepackage{array}
\usepackage{adjustbox}
\usepackage[ruled]{algorithm2e}
\usepackage{amssymb}
\usepackage{amsmath}
\usepackage{arydshln}
\usepackage{multirow}
\usepackage{booktabs}
\usepackage{color}
\usepackage[switch]{lineno}

\usepackage[T1]{fontenc}

\usepackage[utf8]{inputenc}

\usepackage{microtype}

\title{Accelerating Real-Time Question Answering via Question Generation}

\author{Yuwei Fang$^1$ \quad Shuohang Wang$^1$ \quad Zhe Gan$^1$ \quad Siqi Sun$^1$ \\  
\textbf{Jingjing Liu}$^2$ \quad \textbf{Chenguang Zhu}$^1$ \\ \\
     $^1$Microsoft Cognitive Services Research Group, $^2$Tsinghua University \\
    {\small \{\tt yuwfan, shuowa, zhe.gan, siqi.sun, chezhu\}@microsoft.com} \\
    \small \tt JJLiu@air.tsinghua.edu.cn
    }

\begin{document}

\maketitle

\begin{abstract}
Although deep neural networks have achieved tremendous success for question answering (QA), they are still suffering from heavy computational and energy cost for real product deployment.
Further, existing QA systems are bottlenecked by the encoding time of real-time questions with neural networks, thus suffering from detectable latency in deployment for large-volume traffic. 
To reduce the computational cost and accelerate real-time question answering (RTQA) for practical usage, we propose to remove all the neural networks from online QA systems,
and present Ocean-Q (an \textbf{Ocean} of \textbf{Q}uestions),
which introduces a new question generation (QG) model to generate a large pool of QA pairs offline, then in real time matches an input question with the candidate QA pool to predict the answer without question encoding. 
Ocean-Q can be readily deployed in existing distributed database systems or search engine for large-scale query usage, and much greener with no additional cost for maintaining large neural networks.
Experiments on SQuAD(-open) and HotpotQA benchmarks demonstrate that Ocean-Q is able to accelerate the fastest state-of-the-art RTQA system by 4X times, with only a 3+\% accuracy drop.
\end{abstract}

\section{Introduction}
Real-time question answering (QA)~\cite{seo2018phrase,denspi} has enjoyed steadily growing attention in the community due to its commercial value in web-scale applications.
For practical use, QA search engines are required to return the answer given any real-time question within milliseconds in order to avoid human-detectable latency. So far, state-of-the-art QA models~\cite{devlin2018bert,liu2019roberta} rely heavily on cross-sequence attention between a given question and its relevant context in the document. Whenever a new question comes in, it needs to be encoded on the fly and attended to a list of relevant documents, which leads to considerable computational overhead especially when the given context is unlimited (\emph{e.g.}, open-domain setting).

To speed up the process, query-agnostic methods have been proposed to reuse question-related context for different questions.
\citet{seo2018phrase} proposed to build representations of key phrases only in the document and match them with the question representation to obtain answer.
\citet{denspi} further improved the quality of phrase embeddings and proposed a more effective way for caching.
However, the embeddings of incoming questions cannot be cached offline, which still leads to noticeable latency and additional cost of maintaining large neural networks in real applications. 

How can we efficiently reduce the energy cost while simultaneously speeding up real-time question answering for practical use? Here, we propose \textbf{Ocean-Q} (an \textbf{Ocean} of \textbf{Q}uestions), which removes the need of maintaining neural networks for real-time inference, and  significantly reduces question-encoding time in query-agnostic QA methods by leveraging high-quality offline question generation. The proposed pipeline consists of three steps: ($i$) Candidate Answer Extraction from the context; ($ii$) Question Generation based on the candidate answers; and ($iii$) QA-pair Verification by state-of-the-art QA models. Specifically, a candidate pool of all possible QA pairs is generated offline from the data corpus, and when a real-time question comes in, it is directly matched with the candidate pool via $n$-gram overlap to select the answer from the highest-ranked QA pair. 

\begin{figure*}[t!]
\centering
{\includegraphics[width=\linewidth]{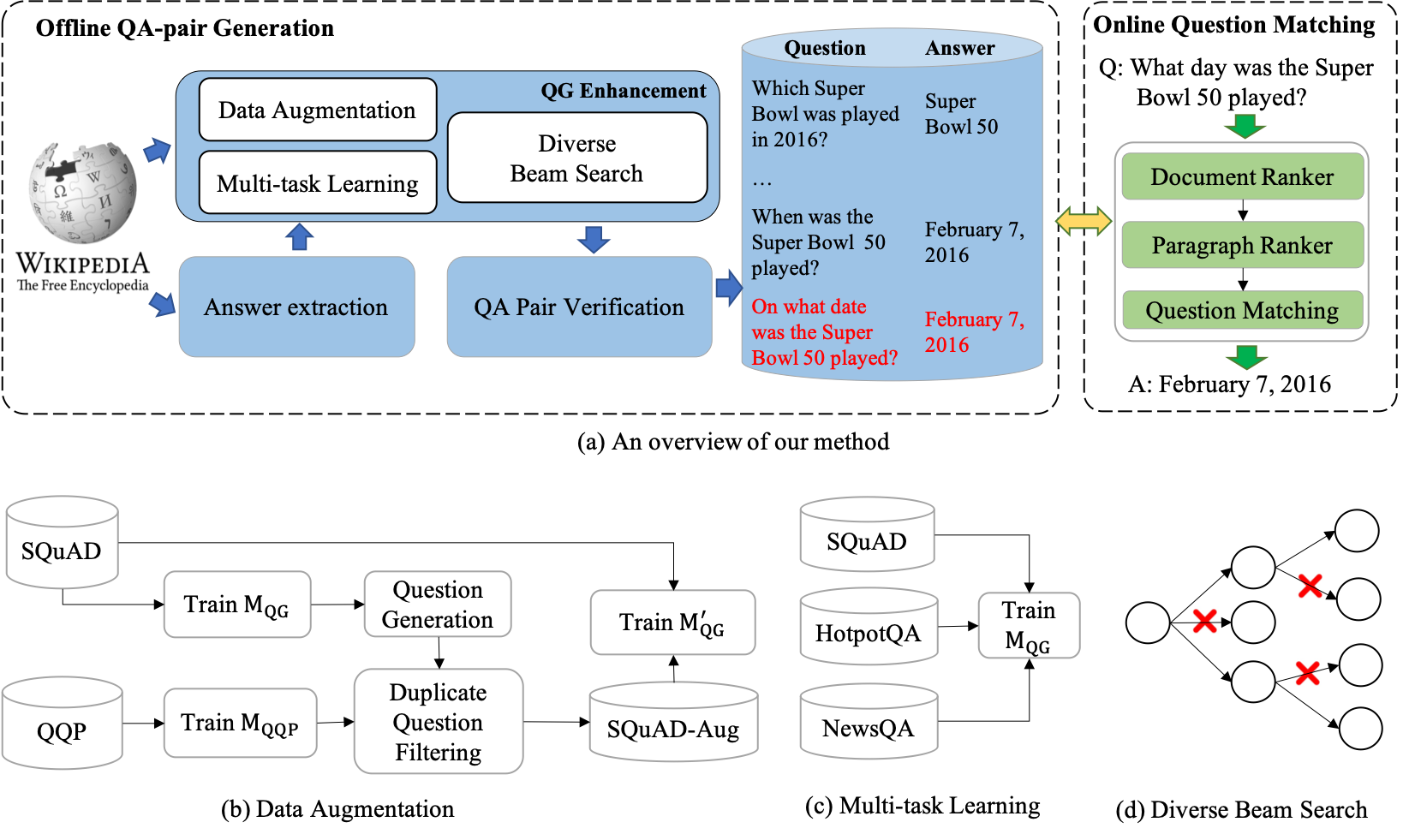}}
\caption{\label{fig:overview} (a) is an overview of the Ocean-Q framework for real-time question answering, consisting of two components: Offline QA-pair Generation and Online Question Matching. Given a set of document paragraphs, steps of Answer Extraction, Question Generation and QA-pair Verification are applied sequentially to generate candidate QA pairs, which are cached into question ``ocean''. To efficiently search over the entire ocean, a two-step ranker is used during online question matching, and the question is then compared with the shortlist of QA pairs via instant $n$-gram matching to derive the right answer. Figures (b), (c) and (d) are used to further enhance Question Generation. Diverse Beam Search will penalize the node extended from the same parent node. QQP: Quora Question Pair dataset for duplicate question detection. SQuAD, HotpotQA, NewsQA: answer-extraction-based question answering datasets.}
\vspace{-10pt}
\end{figure*}

To encourage more organic integration of Question Generation (QG)~\cite{rus2008question, du2017learning, zhang-bansal-2019-addressing, unilm} into QA systems, besides the QG evaluation metrics of BLEU~\cite{papineni2002bleu}, ROUGE~\cite{lin-2004-rouge} and METEOR~\cite{lavie2009meteor}, we also consider RTQA as a new quality criterion for QG models.
The symbiotical connection of QG and RTQA ensures end-to-end performance boost. In addition, we propose a new data augmentation approach and leverage multi-task learning and diverse beam search for QG enhancement to further improve RTQA performance.







Our main contributions are summarized as follows. 
($i$) We set up the first benchmark under the setting of answering open-domain questions without neural networks in real time. To the best of our knowledge, no previous work has leveraged question generation models to reduce the computational cost and mitigate the speed bottleneck in Real-time QA.
($ii$) By generating possible candidate questions offline and circumventing real-time question encoding, Ocean-Q can successfully accelerate existing fastest RTQA system~\cite{denspi} by 4 times, with minimum accuracy drop, and make QA system greener by removing the cost of maintaining and running neural encoders. 
($iii$) We introduce a new data augmentation method and propose the use of multi-task learning and diverse beam search for QG quality enhancement that further boosts RTQA performance. 
($iv$) Experiments show that our proposed approach achieves multifold speedup on generic QA models without sacrificing accuracy much. Our augmented QG component also achieves state-of-the-art performance on QG benchmarks of HotpotQA. 

\section{Related Work}
\paragraph{Open-Domain Question Answering}
Open-domain QA aims to answer any given factoid question without knowing the target domain. 
We focus on text-based open-domain QA system~\cite{chen2017reading}, which first uses information retrieval (IR) to select passages, then applies a machine reading comprehension (MRC) model to extract answers from the passage pool.
Most works rely heavily on cross-sequence attention between questions and related context~\cite{yang2019end,raison2018weaver,wang-etal-2019-multi}, which translates to heavy online computational cost, since every related context needs to be processed repeatedly for a new input question. 
In order to reuse the context, \citet{seo2018phrase} proposed the phrase index QA task by encoding and indexing every possible text span offline into a dense vector.
DENSPI~\cite{denspi} further improved contextualized phrase embeddings with sparse vector, and designed effective caching strategies for real-time QA. 
However, the shared bottleneck of all these works is the demanding time for encoding questions in real time execution.
In this paper, we propose to make full use of a QG-based RTQA framework that is able to achieve 4X speed-up without losing much accuracy.
\vspace{-5pt}

\paragraph{Question Generation}
Question generation was first proposed by \citet{rus2008question} as a task to automatically generate questions from given input. 
Since then, early work mainly focused on using manually designed rules or templates to transform input text to questions~\cite{heilman2011automatic, chali2012towards, lindberg2013generating}.
Due to the recent progress in deep learning, QG research started to adopt the prevailing encoder-decoder framework~\cite{serban-etal-2016-generating,du2017learning,du-cardie-2018-harvesting, liu2019learning, song2018leveraging, zhao-etal-2018-paragraph}.
More recently, large-scale language model pre-training has been applied to QG~\cite{unilm, unilmv2, xiao2020ernie}, which achieved state-of-the-art performance with significant improvement. 
Our QG method is based on the released UniLM model~\cite{unilm}, 
further improving it via novel data augmentation, multi-task learning and diverse beam search.
Moreover, to the best of our knowledge, we are the first work to connect neural question generator with real-time question answering.
\vspace{-5pt}

\paragraph{Green NLP} The primary focus of green NLP is to measure or reduce the amount of required computation to build effective models.
Green AI~\cite{schwartz2019green} covers a broader, long-standing interest in environmentally-friendly scientific research. With the success of pre-trained models~\cite{devlin2018bert} and large-scale datasets~\cite{deng2009imagenet,liu2019roberta}, efficient training and inference becomes increasingly critical regarding the green purpose. 
However, training time is relatively fixed comparing to real large-scale inference requirement, such as queries in Google search.
In this paper, we focus on addressing the green NLP problem regarding the inference efficiency of question answering systems. 
Model distillation~\cite{sun2019patient} or answer-embedding index~\cite{denspi} based methods are widely adopted to speed up the inference time.
However, these works still suffer from maintaining a large neural network to encode questions.
Instead, we propose to remove all the neural networks from the QA system. 
Thus, there will be little computational cost of answering questions, but only some information retrieval cost which can be quite efficient by reverted index. 

\section{Ocean-Q for Real-Time QA}
\label{sec:RTQA}

\subsection{Framework Overview}
\label{subsec:definition}
Our proposed approach to real-time question answering (RTQA) utilises QA pairs generated from a large-scale textual corpus (e.g., Wikipedia). 
The overview of the framework is illustrated in Figure~\ref{fig:overview}, consisting of two main components: ($i$) Offline QA-pair generation; and ($ii$) Online question matching. 

The QA-pair generation pipeline mainly consists of three steps: ($i$) Candidate answer extraction; ($ii$) Answer-aware question generation; and ($iii$) QA-pair verification.
First, we train an answer extraction model to create $K$ possible answer candidates $\{a_{i,j,k}\}_{k=0}^{K-1}$ from paragraph $p_{i,j}$ in document $d_i$. 
Then, for each candidate answer $a_{i,j,k}$, we generate $L$ diverse questions $\{q_{i,j,k,l}\}_{l=0}^{L-1}$ by combining $p_{i,j}$ and $a_{i,j,k}$ as the input of an answer-aware QG model. 
To further validate the generated QA pairs, a BERT-based QA model is utilized for authentication. 
By these three steps, we obtain approximately $N*M*K*L$ QA pairs from the whole Wikipedia, where $N$ is the number of documents, $M$ is the average number of paragraphs in a document, $K$ is the average number of answer candidates in a paragraph, and $L$ is the number of questions for each answer candidate. 
With these pairs, the RTQA task is formulated as a ranking function $R$, where given an input question $\tilde{q}$, the goal is to predict the answer as $a=\arg\max_{i,j,k,l} R(q_{i,j,k,l}, \tilde{q})$.

\subsection{Offline QA-pair Generation}
\paragraph{Answer Extraction}
In principle, any span in a paragraph could be an answer candidate. 
To maintain high recall for QA, we formulate answer extraction as a classification problem and use pre-trained RoBERTa~\cite{liu2019roberta}.
For each token in the paragraph, we predict its probability of being the start and end position of the answer. Then, predicted answers with top-$K$ scores are selected as candidate answers.
Assume the encoded representation of a paragraph is denoted as  $\mathbf{C} \in \mathbb{R}^{n \times d}$, where $n$ is the length of the paragraph and $d$ is the dimension of each contextualized token representation. Then, the logits of every position being the start and end of the candidate answer span can be computed by a linear layer on top of $\mathbf{C}$:
\begin{align}
    \mathbf{o}_{s} = \mathbf{C}\mathbf{w}_{1},\,\,
    \mathbf{o}_{e} = \mathbf{C}\mathbf{w}_{2} \,,
\end{align}
where $\mathbf{o}_{s},\mathbf{o}_{e}\in \mathbb{R}^{n}$ and $\mathbf{w}_{1},\mathbf{w}_{2}\in \mathbb{R}^{d}$.
During training, the ground-truth distribution of a position being the start and end of an answer span is calculated over all possible answers in each paragraph, denoted as $P_s$ and $P_e$. 
A KL divergence loss is then defined as:
\begin{equation}
    \mathcal{L} = D_{KL}(P_{s}||Q_{s}) +  D_{KL}(P_{e}||Q_{e})\,,
\end{equation}
where $Q_s$ and $Q_e$ are the predicted distributions of a position being the start and end of an answer span based on the logits $\mathbf{o}_{s}$ and $\mathbf{o}_{e}$.
During inference, we first calculate the start and end logits for each token in the paragraph, then top-$K$ answer spans are extracted based on the score below:
\begin{equation}
    s_k = \mathbf{o}_{s}[a_{s,k}] + \mathbf{o}_{e}[a_{e,k}]\,,
\end{equation}
where $s_k$ is the score for the $k$-th answer candidate $a_k$ with start and end position being $a_{s,k}$ and $a_{e,k}$.


\paragraph{Question Generation}
For question generation, we build upon the released state-of-the-art UniLM model~\cite{unilm}. 
To generate $L$ different questions given an answer span, we adopt standard beam search strategy.
More formally, we concatenate a given input passage $p_{i,j}$ and an answer span $a_{i,j,k}$ as the first segment (denoted as $X$), and incrementally generate each token of the target question.
At time step $t-1$, the model keeps track of $L$ hypotheses and their scores $S(q_{t-1}|X)$, where $L$ is the beam size and $q_{t-1}$ is a generated token. 
Then, at time step $t$, each hypothesis selects $L$ extensions, which leads to $L*L$ hypotheses in total.
In the end, the top-$L$ generated questions are selected from $L*L$ hypotheses at the final time-step $T$. We further explore different strategies to improve QG model, detailed in the Enhanced Question Generation Section. 

\paragraph{QA-pair Verification}
With the aforementioned two steps, we can obtain an initial set of QA pairs.
However, these answers may not be accurate enough since answer extraction model could not precisely know where is the answer boundary.
For example, the extracted span ``February 7 '' might be a sub span of the gold answer ``February 7, 2016'' which leads to a lower RTQA performance.
Therefore, we use the initial QA dataset to train a BERT-based QA model, to validate the initial QA pool.
Then, we replace the the initial answer $a_{i,j,k}$ with the predicted answer $a_{i,j,k}^{\prime}$ from the QA model.

\subsection{Online Question Matching}
The main challenge in real-time question matching is how to efficiently search over the entire question ``ocean'' that consists of $N*M*K*L$ questions. 
Given that the ranking function $R$ can be a simple scoring function, the problem can be redefined as finding accurate paragraphs among $N*M$ paragraphs that may lead to the correct answer. Therefore, we propose a two-step ranker and conduct question matching as follows.

\paragraph{A Two-step Ranker}
DrQA~\cite{chen2017reading} has demonstrated strong capability in finding relevant documents based on bi-gram hashing and TF-IDF matching.
Therefore, we leverage DrQA to build a document-level TF-IDF sparse vector $\mathbf{d}_i$ for each document. 
The input question is also encoded into a bi-gram TF-IDF sparse vector $\mathbf{q}$. 
Then, top-$n$ documents are retrieved from $N$ documents by calculating the inner product between $\mathbf{q}$ and each $\mathbf{d}_i$.
Similarly, we also compute a second paragraph-level TF-IDF $\mathbf{p}_i$ for each paragraph after the first ranker, and collect top-$m$ paragraphs only based on the inner product score on paragraph-level.
Since paragraphs capture more local information than documents, by implementing the two-step ranker, we can safely reduce the number of paragraphs from $N*M$ to $m$ ($m \ll N*M$) for question ranking in the final step.

\paragraph{Question Matching}
After the proposed two-step ranking process, the total number of candidate questions reduces to $m*K*L$ (around 200k, if $m,K,L$ are $100, 100, 20$, respectively).
We simply calculate the common number of unique tokens to measure the similarity between each question candidate $q_i$ and the input question $\tilde{q}$. Specifically,
\begin{equation}
    score(\tilde{q}, q_i) = \frac{\tilde{\mathbf{q}}\cdot \mathbf{q}_i}{|\tilde{\mathbf{q}}| + |\mathbf{q}_i|}\,,
\end{equation}
where $\tilde{\mathbf{q}}$ and $\mathbf{q}_i$ are binary sparse representations, indicating whether each token in the vocabulary exists in the question or not. $|\tilde{\mathbf{q}}|$ and $|\mathbf{q}_i|$ are the $\ell_1$ norm of each vector. After question matching, we use the corresponding answer to the retrieved question as the final predicted answer. 

\subsection{Enhanced Question Generation}
\label{sec:qg_eval}
Question Generation is a critical component in Ocean-Q framework. We hypothesize that more diverse questions from QG model may enhance the RTQA performance.
To verify this, we investigate from three perspectives: ($i$) we propose a new data augmentation method to increase in-domain training data diversity; ($ii$) we leverage multi-task learning with out-of-domain training data; ($iii$) we leverage diverse beam search in decoding phase.

\paragraph{Data Augmentation}
To increase the diversity of the QG training dataset $D_1$, we select additional samples from the generated question pool and reuse them as training data. 
A caveat is that the generated questions may semantically diverge from the original question, so we aim to select only a small portion from the pool to augment the training data. 
To this end, we train a question paraphrasing model on the Quora Question Pairs (QQP) dataset\footnote{https://www.quora.com/q/quoradata/First-Quora-Dataset-Release-Question-Pairs} $D_2$ to obtain a paraphrase score between the generated question and the ground-truth question, and use it as one of the guidelines for question selection.
Moreover, since semantically-similar questions may overlap with each other, we also use the $F_1$ metric to calculate the number of overlapping tokens between the generated question and the ground-truth question as another guideline.
Thus, for each data sample in $D_1$, we can effectively craft augmented training samples, forming a new dataset $D_3$, and the final QG model $M_{QG}^{\prime}$ is trained by combining $D_1$ and $D_3$. 
The proposed data augmentation method is summarized in Algorithm~\ref{alg:DataAugmentation}.

\paragraph{Multi-task Learning}
Multi-task learning (MTL) \cite{caruana1997multitask} has been proved to be effective for improving model generalization.
However, directly mixing all the tasks has the potential downside of biasing the model towards auxiliary tasks instead of the intended target task. 
Therefore, \citet{xu-etal-2019-multi} proposed a mixed ratio strategy and verified its effectiveness in MRC tasks.
Inspired by this, we apply the MTL algorithm to QG:
given $K$ different tasks and $D_1$ as the target task, we use all mini-batches from $D_1$ and sample specified ratio of mini-batches from the remaining tasks.
In experiments, we use hyper-parameter search to find the best ratio for each target task.

\begin{algorithm}[t!]
\small
\caption{\label{alg:DataAugmentation}Data Augmentation for QG.}
\SetAlgoNoLine
\SetArgSty{}
\KwIn{QG dataset $D_1$, QQP dataset $D_2$, QQP score threshold $s_{min}, s_{max}$, $F_1$ score threshold $f_{min}, f_{max}$, maximum selection number $N$}
 Train QG model $M_{QG}$ on $D_1$\;
 Train QQP model $M_{QQP}$ on $D_2$\;
 $D_3$ = \{\}\;
 \ForEach{$ \langle p,q,a \rangle \in D_1$}{
    Generate $L$ different questions with beam search 
    $q_1, .., q_L = M_{QG}(p,a)$\;
    Calculate QQP score for each question
    $s_{qqp}^{i}=M_{QQP}(q_i, q), 1 \le i \le L$\;
    Calculate $F_1$ score for each question
    $s_{f1}^{i}=F_1(q_i, q), 1 \le i \le L$\;
    $D_{sel}$ = \{\}\;
    \For{$i=1$ \KwTo $L$}{
        \If{$s_{min} \le s_{qqp}^{i} \le s_{max} $ and 
            $f_{min} \le s_{f1}^{i} \le f_{max} $ and 
            $len(D_{sel}) \le N$}{
        add $\langle p,q_i,a \rangle$ to $D_{sel}$
        }
    }
    add $D_{sel}$ to $D_3$ \;
  }
  Train QG model $M_{QG}^{\prime}$ on $D_1$ and $D_3$ \;
\KwOut{$M_{QG}^{\prime}$}
\end{algorithm}

\paragraph{Diverse Beam Search}
Standard beam search tends to generate similar sequences in the $n$-best list. 
To mitigate this, \citet{li2016simple} proposed to penalize the hypotheses that are extended from the same parent node, effectively favoring outlets from more diverse parents.
In our work, we leverage this directly to generate more diverse questions for each candidate answer.
More formally, the score for each question hypothesis is written as:
\begin{equation}
    S(q_{t-1}^k, q_{t}^{k,k^{\prime}}|X) = S(q_{t-1}^k, q_{t}^{k,k^{\prime}}|X) - {\gamma}k^{\prime}\,,
\end{equation}
where $X$ is the context, $q_{t-1}^k$ is the $(t-1)$-th generated token at beam $k$, $q_{t}^{k,k^{\prime}}$ is the $t$-th generated token at beam $k^\prime$ based on $q_{t-1}^k$, $k^{\prime}$ is the rank of the current hypothesis among its siblings, and $\gamma$ is the diversity rate. $S(q_{t-1}^k, q_{t}^{k,k^{\prime}}|X)$ is the ranking score used in standard beam search. In experiments, we show that using diverse beam search can effectively improve RTQA performance.
\section{Experiments}

\subsection{Experimental Setup}
\paragraph{Datasets} We conduct experiments on SQuAD~\cite{rajpurkar-etal-2016-squad} and HotpotQA~\cite{yang2018hotpotqa} to validate the effectiveness of our proposed approach.
SQuAD is one of the largest QA datasets that requires mostly single-hop reasoning, while HotpotQA is a popular benchmark for multi-hop QA. Both are collected over 100K questions by presenting a set of Wikipedia articles to crowdworkers. 

\paragraph{Metrics} For RTQA evaluation, we follow DENSPI~\cite{denspi} to compare EM, F1 and inference speed (s/Q) on open-domain SQuAD (SQuAD-Open) and SQuAD v1.1 development sets as our test sets. The answer extraction and QA-pair verification models are trained on the original SQuAD v1.1 training set. 
The question generation model is trained on SQuAD, HotpotQA and NewsQA datasets through multi-task learning. 

For QG evaluation, we follow~\citet{du-cardie-2018-harvesting} to split the original training set into training and test on SQuAD, and follow~\citet{pan-etal-2020-DQG} to use their released data split for a fair comparison on HotpotQA.
Besides commonly used metrics such as BLEU-4, METEOR and ROUGE-L, we also use final RTQA performance to measure QG quality.
Since~\citet{pan-etal-2020-DQG} pre-processes the original HotpotQA dataset to select relevant sentences because of the difficulty in producing accurate semantic graphs for very long documents, 
the RTQA score on HotpotQA is based on the evidence sentences rather than the whole context. 

\paragraph{Implementation Details} Our implementation is mainly based on HuggingFace's Transformers~\cite{Wolf2019HuggingFacesTS}, UniLM~\cite{unilm} and DrQA~\cite{chen2017reading}. We conduct experiments on 16 NVIDIA V100 GPUs and Intel(R) Xeon(R) W-2175 CPU.
Each model in Ocean-Q framework took several hours to finetune.
For offline QA-pair generation from Wikipedia, it took about one week. Datasets and more implementation details can be found in the Appendix. 

\subsection{Real-Time QA Experiments}

\begin{table}[t!]
\centering
\resizebox{1.0\columnwidth}{!}{
\begin{tabular}{lcccc}
\toprule
 & Parameters & EM & F1 & s/Q \\
 \midrule
DrQA & 68M & 29.8 & - & 35\\
R$^3$ & - &  29.1 & 37.5 & - \\
Paragraph ranker & 68M &  30.2 & - & - \\
Multi-step reasoner & 68M &  31.9 & 39.2 & - \\
MINIMAL & - &  34.7 & 42.5 & - \\
BERTserini & 110M &  38.6 & 46.1 & 115 \\
Weaver & - & -  & 42.3 &  - \\
DENSPI-Hybrid & 336M & 36.2 & 44.4 & 1.24$^*$ \\ \midrule
Ocean-Q & 0 & 32.7 & 39.4 &  0.28 \\
\bottomrule
\end{tabular}
}
\caption{\label{table:SQuAD-open} Results on SQuAD-open. `s/Q' is the average seconds per query in CPU environment. 
Note that DENSPI does not include the inference time of question embedding in their evaluation. We used their code to load question embedding on the dev set, which cost 0.43 seconds on average and was added to their reported numbers, marked as asterisk (*).}
\label{sec:exp-sta}
\end{table}

\begin{table}[t!]
\centering
\begin{adjustbox}{scale=0.90,center}
\begin{tabular}{cccc}
\toprule
 & Model & EM & F1 \\
 \midrule
\multirow{2}{*}{Original} & DrQA & 69.5 & 78.8 \\
& BERT & 84.1 & 90.9 \\ \midrule
\multirow{3}{*}{\shortstack{Query-\\Agnostic}} & LSTM+SA+ELMo & 52.7 & 62.7 \\ 
& DENSPI  & 73.6 & 81.7 \\ 
& Ocean-Q & 63.0 & 70.5 \\ \bottomrule
\end{tabular}
\end{adjustbox}
\caption{Results on SQuAD v1.1. DrQA~\cite{chen2017reading} and BERT~\cite{devlin2018bert} results are reported by SQuAD leaderboard. LSTM+SA+ELMo~\cite{seo2018phrase} and DENSPI~\cite{denspi} are two query-agnostic baselines.}
\label{table:SQuAD-close}
\vspace{-10pt}
\end{table}
\label{sec:exp-qa}

\paragraph{Baselines} 
On SQuAD-Open, state-of-the-art (SOTA) models include DrQA~\cite{chen2017reading}, R$^3$~\cite{wang2018r}, Paragraph ranker~\cite{lee2018ranking}, Multi-step reasoner~\cite{das2019multi}, MINIMAL~\cite{min2018efficient}, BERTserini~\cite{yang2019end}, Weaver~\cite{raison2018weaver} and DENSPI~\cite{denspi}. 
On SQuAD v1.1, we follow~\citet{denspi} to compare with two types of model. The first group includes DrQA~\cite{chen2017reading} and BERT~\cite{devlin2018bert} that rely on cross attention between the question and context. 
The second group includes LSTM+SA+ELMO~\cite{seo2018phrase} and DENSPI~\cite{denspi} as query-agnostic models. 
The accuracy comparison in following experiments is based on subtracting the EMs from two comparing systems.

\paragraph{Results}
Table~\ref{table:SQuAD-open} and~\ref{table:SQuAD-close} summarize results on SQuAD-Open and SQuAD v1.1 datasets.
In open domain setting:
$(i)$ Ocean-Q outperforms DrQA by 3\% and achieves 125x speedup.
This indicates that Ocean-Q could outperform earlier cross-attention model while achieving much faster inference speed.
$(ii)$ Compared to BERTserini, Ocean-Q achieves 410 times speedup while sacrificing accuracy by 5.8\%.
As pointed out in~\citet{denspi}, the main gap between query-agnostic model and cross-sequence attention model is due to the \emph{decomposability gap}.
$(iii)$ Ocean-Q accelerates DENSPI by 4 times at the expense of 3.5\% accuracy drop.
We call this gap the \emph{generative gap}, since Ocean-Q relies on question generation model to enumerate possible questions for RTQA. A more rigorous QG model design to generate more diverse questions may help close this gap.
In close-domain setting (a simpler problem than open domain), Table~\ref{table:SQuAD-close} shows that Ocean-Q outperforms the query-agnostic baseline~\cite{seo2018phrase}, which drops 30.3\% from close- to open-domain; while DENSPI, BERT and DrQA drop 37.4\%, 45.5\%, and 39.7\%, respectively.
This demonstrates Ocean-Q's better generalization ability from close- to open-domain.

\subsection{Question Generation Experiments}
\label{subsec:qg-exp}

\begin{table}[t!]
\centering
\resizebox{1.0\columnwidth}{!}{
\begin{tabular}{lccccc}
\toprule
& \multicolumn{3}{c}{QG} & \multicolumn{2}{c}{RTQA} \\
& BLEU-4 & MTR & RG-L & EM & F1 \\ \midrule
\multicolumn{6}{l}{\emph{SQuAD}} \\ \midrule
CorefNQG & 15.16 & 19.12 & - & - & - \\
SemQG & 18.37 & 22.65 & 46.68 & - & - \\
UniLM & 22.78 & 25.49 & 51.57 & - & - \\
UniLMv2 & 24.70 & 26.33 & 52.13 & - & - \\
Our baseline & 22.89 & 25.46 & 51.52 & 62.33 & 70.09 \\
+ DA-sim & 23.72 & 26.07 & 52.23 & 62.03 & 69.68 \\
+ DA-div & 22.15 & 25.13 & 51.29 & 62.71 & 70.34 \\
+ MTL & 23.16 & 25.74 & 51.94 & 63.04 & 70.52 \\
+ DBS & 22.89 & 25.46 & 51.52 & 62.43 & 70.26  \\ \midrule
\multicolumn{6}{l}{\emph{HotpotQA}} \\ \midrule
SRL-Graph & 15.03 & 19.73 & 36.24 & - & - \\
DP-Graph & 15.53 & 20.15 & 36.94 & - & - \\
Our baseline & 26.46 & 26.75 & 46.64 & 42.93 & 54.18 \\
+ DA-sim & 28.53 & 28.36 & 48.78 & 44.39 & 55.79 \\
+ DA-div & 26.94 & 27.27 & 47.18 & 44.19 & 55.43 \\ 
+ MTL & 26.76 & 27.03 & 46.97 & 44.30 & 55.73 \\
+ DBS & 26.46 & 26.75 & 46.64 & 43.53 & 54.91 \\
\bottomrule
\end{tabular}
}
\caption{\label{table:QG-QA} Results of QG Enhancement on SQuAD and HotpotQA. The first group follows the data split in~\citet{du-cardie-2018-harvesting} on SQuAD, while the second group follows~\citet{pan-etal-2020-DQG} on HotpotQA. MTR is short for METEOR, and RG-L for ROUGE-L.}
\vspace{-10pt}
\end{table}

\begin{table*}[t!]
\centering
\begin{small}
\begin{tabular}{lll}
\toprule
\textbf{Passage} & \multicolumn{2}{p{0.85\textwidth}}{\textcolor{blue}{Super Bowl 50} was an American football game to determine ... The \textcolor{blue}{American Football Conference (AFC)} champion \textcolor{blue}{Denver Broncos} defeated the \textcolor{blue}{National Football Conference (NFC)} champion \underline{\textcolor{blue}{Carolina Panthers 24-10}} to earn their third \textcolor{blue}{Super Bowl} title. The game was played on ... }\\
\textbf{QA Pairs} & What was the winning score in the Super Bowl? & 24-10 \\
& \textcolor{red}{What was the final score of the Super Bowl?} &  \textcolor{red}{24-10} \\
& Who did the Denver Broncos defeat in the Super Bowl? & Carolina Panthers \\
\textbf{Question} & What was the final score of Super Bowl 50? & \\ 
\midrule
\textbf{Passage} & \multicolumn{2}{p{0.85\textwidth}}{Demographically, it was the most diverse city in \textcolor{blue}{Poland}, with significant numbers of \textcolor{blue}{foreign-born inhabitants}. … In \textcolor{blue}{1933}, out of \underline{\textcolor{blue}{1,178,914}} inhabitants \textcolor{blue}{833,500} were of \textcolor{blue}{Polish mother tongue}. \textcolor{blue}{World War II} changed the demographics of the city, and to this day ...} \\
\textbf{QA Pairs} & What was the population of Warsaw in 1933? & 1,178,914 \\
& How many people in 1933 had Polish mother tongue? & 833,500 \\
& \textcolor{red}{How many inhabitants in 1933 had Polish mother tongue?} & \textcolor{red}{833,500} \\
\textbf{Question} & How many of Warsaw's inhabitants spoke Polish in 1933? & \\
\bottomrule
\end{tabular}
\end{small}
\caption{Qualitative examples of generated QA pairs and predictions. {\color{blue}Blue} and {\color{red} red} represent the extracted answers and matched QA pair. Sampled QA pairs generated by Ocean-Q correspond to the \underline{\textcolor{blue}{underline}} answer span.}
\label{tab:examples}
\vspace{-5pt}
\end{table*}

\paragraph{Baselines}
On SQuAD, we compare with SOTA baselines: CorefNQG~\cite{du-cardie-2018-harvesting}, SemQG~\cite{zhang-bansal-2019-addressing}, UniLM~\cite{unilm} and UniLMv2~\cite{unilmv2}. 
On HotpotQA, our chosen baselines are SRL-Graph~\cite{pan-etal-2020-DQG} and DP-Graph~\cite{pan-etal-2020-DQG}. 
We reproduced the results from the released SOTA model UniLM and used this as our baseline.

\paragraph{Results}
Table~\ref{table:QG-QA} compares Ocean-Q with baselines over QG metrics as well as final RTQA scores.
For data augmentation (DA), We use two different F1 thresholds. One from 0.5 to 1.0 (DA-sim) is intended to select similar questions, while the other one from 0.0 to 0.5 (DA-div) tends to select more diverse questions.
Our observations from the comparison results:
$(i)$ DA-div, multi-task learning (MTL) and diverse beam search (DBS) improve RTQA score by 0.38/0.71/0.10 and 1.26/1.37/0.60 points, respectively, over the baseline on SQuAD and HotpotQA.
This indicates that diversified questions could be beneficial for lifting RTQA performance.
$(ii)$ DA improves 0.83/0.61/0.71 and 2.07/1.61/2.14 points, respectively, over our strong baseline on SQuAD and HotpotQA.
Compared to current SOTA, it still falls behind by 0.98/0.26/0.10 points over UniLMv2 on SQuAD, but it significantly boosts 13.0/8.21/11.84 points over DP-Graph on HotpotQA.
$(iii)$ MTL improves the most for RTQA, and DA improves the most on QG metrics.
This indicates that RTQA benefits more from out-of-domain data, while QG metrics gain more from in-domain data. 
$(iv)$ Similar in-domain data help more on QG metrics than diverse data.
Comparing DA-sim with our baseline, it improves 0.83/0.61/0.71 and 2.07/1.61/2.14 points on SQuAD and HotpotQA, respectively. In contrast, DA-div is worse than the baseline on SQuAD and achieves less gain on HotpotQA.

\begin{figure}[t!]
\centering
{\includegraphics[width=\linewidth]{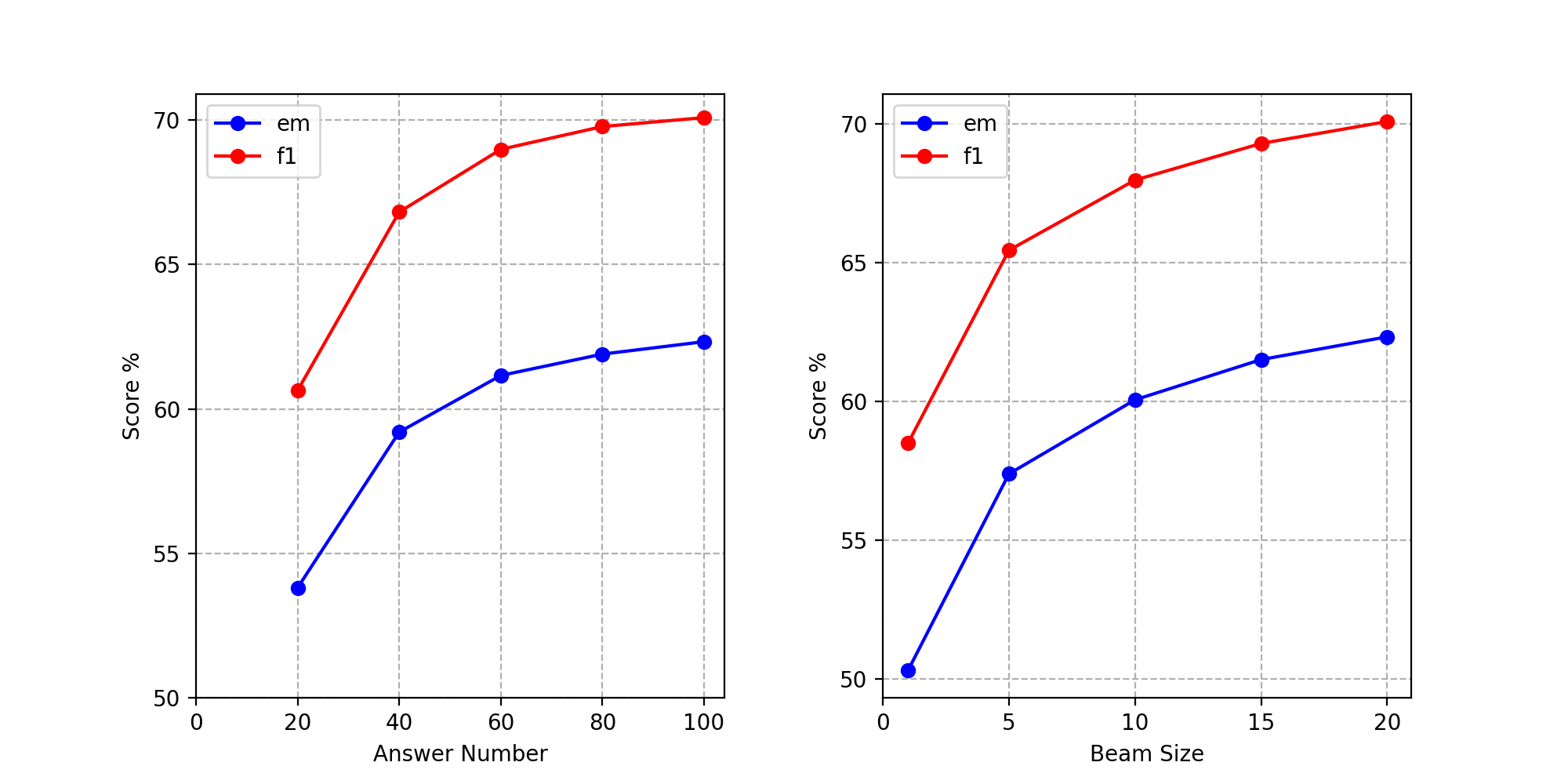}}
\caption{\label{fig:effects_ans_beam} Effects of increasing the number of candidate answers and beam size on SQuAD RTQA performance. When answer number is 100 and beam size is 20, the setting degenerates to our baseline in Table~\ref{table:QG-QA}.}
\vspace{-10pt}
\end{figure}



\begin{table}[t!]
\centering
\begin{adjustbox}{scale=0.90,center}
\begin{tabular}{cccc}
\toprule
Retriever & \#Docs & \#Paras & Accuracy(\%) \\
 \midrule
DrQA & 5 & 214 & 77.8 \\ \midrule
\multirow{3}{*}{\shortstack{Two-step\\Ranker}} & 20 & 100 & 80.6 \\
& 20 & 150 & 82.4 \\
& 20 & 200 & 83.4 \\
\bottomrule
\end{tabular}
\end{adjustbox}
\caption{\label{table:eff_ranker}Paragraph retrieval results on SQuAD-Open.
`\#Docs' and `\#Paras' are the number of retrieved document and paragraph per query. `Accuracy' is the percentage of questions whose answers appear in the retrieved paragraphs. DrQA~\cite{chen2017reading} retrieves 5 documents for each question. 
For the two-step ranker, we retrieve 100, 150 and 200 paragraphs from 20 Wikipedia documents for each question.}
\end{table}

\subsection{Analysis}
We provide four analysis to evaluate the effectiveness of each component in Ocean-Q framework: $(i)$ Effect of different numbers of candidates in answer extraction; $(ii)$ Effect of beam size in question generation; $(iii)$ Ablation on QA pair verification; and (iv) Effect of the two-step ranker in online question matching. 
Figure~\ref{fig:effects_ans_beam} show that RTQA score receives consistent improvement by increasing the number of candidate answers and beam search size, since both methods generate more diverse candidate question-answer pairs. By increasing the answer number from 10 to 100, RTQA score receives 16.96/18.85 points improvement. By enlarging beam size from 1 to 20, there is 12.02/11.59 points improvement. 
For QA-pair verification, we see 9.46\% accuracy drop if ablating this step in the experiment, which demonstrates that QA pair verification is useful for improving RTQA performance.
For two-step ranker, Table~\ref{table:eff_ranker} shows that by using the ranker, our model achieves 5.6\% higher accuracy than DrQA using similar number of paragraphs (200 vs. 214), and 2.8\% higher using just half (100 vs. 214).

\paragraph{Qualitative Analysis}
Table~\ref{tab:examples} shows two qualitative examples of generated QA pairs and predictions by Ocean-Q. 
In the first example, the extracted answer span ``Carolina Panthers 24-10'' includes two possible sub answers ``Carolina Panthers'' and ``24-10''.
This phenomena is expected since the answer extraction model cannot precisely know where is the answer boundary.
However, QG model can generate different questions through attention over different sub answers, and the QA verification model can further correct inaccurate answers from the answer extraction phase.
In the second example, we note that the same input answer span ``1,178,914'' generates questions that correspond to different spans in the passage. 
We hypothesize that since ``1,178,914'' and ``833,500'' appear close to each other in the passage, QG model may not distinguish them from the context and thus generate inaccurate questions.
But the QA verification model can help correcting the false answer.

\paragraph{Green NLP Analysis} As shown in Table~\ref{sec:exp-sta}, we are the first method that can answer questions without using neural networks. 
Usually, models with more parameters can achieve better performance, especially when they are initialized by pre-trained models.
Our model without neural networks during inference can achieve very competitive performance.
Compared to the baseline of DENSPI, we both need to save a large amount of answer candidates into disk. 
However, DENSPI still needs to compute the question embedding whenever a new question comes in.
Our method does not require a heavy computational cost from neural networks, but only the information retrieval cost based on inverted index which is included in all the baseline methods.
When deployed into large-scale question answering systems, such as Google, our method can be greener than DENSPI.

\section{Conclusion}
In this paper, we present Ocean-Q, a novel method that uses QG to reduce the computational cost of RTQA, and address the bottleneck of encoding coming question which cannot be cached offline. 
Experiments on SQuAD-Open demonstrate that Ocean-Q is able to accelerate the fastest RTQA system by 4 times, while only losing the accuracy by 3+\%.
To further improve QG quality, we introduce a new data augmentation method and leverage multi-task learning and diverse beam search to boost RTQA performance.
Currently, there still exists a gap between Ocean-Q and the state-of-the-art RTQA models. 
Future work that includes designing a better QG model may help closing the gap.

\clearpage

\bibliography{anthology,custom}
\bibliographystyle{acl_natbib}

\clearpage
\appendix
\section{Datasets}
\label{appendix:dataset}
\paragraph{QG} On SQuAD, we use preprocessed dataset\footnote{\url{https://drive.google.com/open?id=11E3Ij-ctbRUTIQjueresZpoVzLMPlVUZ}} from~\citet{unilm} to fair compare with previous works~\cite{du-cardie-2018-harvesting, zhang-bansal-2019-addressing, unilm, unilmv2}. 
As the test set is hidden, they split the original training set into 75,722 training and 11,877 test samples. The original development set that includes 10,570 samples is used for validation. On HotpotQA, we use preprocessed dataset\footnote{\url{https://drive.google.com/drive/folders/1vj2pWYZ7s08f4qP8vHAd3FEHnYa_CTnn}} from~\citet{pan-etal-2020-DQG} which includes 90,440 / 6,972 examples for training and evaluation, respectively. For multitask-learning, NewsQA data\footnote{\url{https://github.com/mrqa/MRQA-Shared-Task-2019}} includes 74,159 training samples.

\paragraph{RTQA} We use SQuAD v1.1\footnote{\url{https://github.com/rajpurkar/SQuAD-explorer/blob/master/dataset}} for RTQA training and testing. Answer extraction and QA pair verification models are trained on the whole training set that includes 87,599 examples. Follow previous work, the development set is used as the test set since the official test set is hidden. For HotpotQA, we exclude questions whose answers are ``yes/no'' as they does not appear in the passage and this lead to 82,095 / 6,515 examples for training and testing respectively.

\section{Implementation Details}
\label{appendix:implementation}

\paragraph{Answer Extraction}
We use pre-trained RoBERTa-large model (355M parameters) and finetune on the original SQuAD training set for 3 epochs, with batch size 8, max sequence length 512 and learning rate 1e-5. 
During inference, we limit the maximum answer length to 10 and extract 100 candidate answers per paragraph.

\paragraph{Question Generation}
We use pre-trained UniLM-large model (340M parameters) and their default parameters.
Specifically, we finetune on the training set for 10 epochs, with batch size 32, masking probability 0.7, learning rate 2e-5, and label smoothing 0.1.
For QQP model, we use HuggingFace's transformers~\cite{Wolf2019HuggingFacesTS} and finetune on QQP dataset for 3 epochs, with batch size 16, gradient accumulation steps 2, max sequence length 256, and learning rate 1e-5.
For data augmentation, $s_{min}$, $s_{max}$ and $N$ are set to 0.5, 1.0 and 2 respectively.
For multi-task learning, we use SQuAD, HotpotQA and NewsQA dataset, with mixture ratio 0.4 by searching from 0.1 to 0.9.
For diverse beam search, the diverse strength rate $\gamma$ is set to 4.0.
For RTQA, we generate 20 different questions with each answer.

\paragraph{QA-pair Verification}
We use pre-trained BERT Whole-Word-Masking model (340M parameters) and finetune on the original SQuAD training set for 4 epochs, with batch size 3, gradient accumulation steps 8, max sequence length 384, doc stride 128 and learning rate 3e-5.

\paragraph{Online Question Matching}
For building document and paragraph TF-IDF vector, we use the default configuration of DrQA which uses bigram TF-IDF and map the vectors to $2^{24}$ bins with an unsigned murmur3 hash.
During online question matching, we retrieve top 100 paragraphs from 20 Wikipedia documents.

\section{Inference Speed Analysis}
\begin{table}[t!]
\centering
\resizebox{1.0\columnwidth}{!}{
\begin{tabular}{lccc}
\toprule
& Doc Ranker & Para Ranker & Question Matching \\ \midrule
s/Q & 0.126 & 0.026 & 0.130 \\
\bottomrule
\end{tabular}
}
\caption{\label{table:rtqa_time} Inference speed for each component in Ocean-Q framework on SQuAD-Open. `s/Q' is the average seconds per query.}
\end{table}

Table~\ref{table:rtqa_time} shows the inference speed for each component in Ocean-Q framework. We note that document ranker and question matching are the main cost.
We expect Ocean-Q can achieve greater speed-up by further optimizing on these two parts.


\end{document}